\documentclass[10pt,a4paper,oneside,onecolumn]{article}

\usepackage{ICROMA}

\usepackage{enumerate}
\usepackage{amsmath,amssymb}
\usepackage{graphicx}
\usepackage{float}
\usepackage{caption}
\usepackage{subcaption}
\usepackage{algorithm}
\usepackage{algpseudocode}
\usepackage{easyReview}

\title{CVCM Track Circuits Pre-emptive Failure Diagnostics for Predictive Maintenance Using Deep Neural Networks}
\author{
        Debdeep Mukherjee\textsuperscript{2, *}, 
        Di Santi Eduardo\textsuperscript{1, *}, 
        Lefebvre Clément\textsuperscript{1, *}, 
        Mijatovic Nenad\textsuperscript{1, *},\\
        Martin Victor\textsuperscript{1, *}, 
        Josse Thierry\textsuperscript{3, *}, 
        Jonathan Brown\textsuperscript{1, *},
        Saiah Kenza\textsuperscript{1, *}
}
\affiliation{
        \textsuperscript{1}Digital and Integrated Systems, Alstom\\
        \textsuperscript{2}Innovation and Smart Mobility, Alstom\\
        \textsuperscript{3}Project System Engineering, Alstom\\
\noindent
        E-mail:\href{mailto:debdeep.mukherjee@alstomgroup.com}{debdeep.mukherjee@alstomgroup.com},
        \href{mailto:eduardo.di-santi@alstomgroup.com}{eduardo.di-santi@alstomgroup.com},
        \href{mailto:clement.lefebvre-renard@alstomgroup.com}{clement.lefebvre-renard@alstomgroup.com},
        \href{mailto:nenad.mijatovic@alstomgroup.com}{nenad.mijatovic@alstomgroup.com},
        \href{mailto:victor-andres.martin@alstomgroup.com}{victor-andres.martin@alstomgroup.com},
        \href{mailto:josse.thierry@alstomgroup.com}{josse.thierry@alstomgroup.com},
        \href{mailto:jonathan.brown@alstomgroup.com}{jonathan.brown@alstomgroup.com},
        \href{mailto:kenza.saiah@alstomgroup.com}{kenza.saiah@alstomgroup.com}
}

\begin{document}

\maketitle

\begin{abstract}
    Track circuits are crucial for railway operations. They are the main signalling sub-system used to locate trains within track segments. Continuous Variable Current Modulation (CVCM) is one of such track circuit technologies. As with any equipment deployed on field, this safety critical component may experience failures, provoking cascading blockages of operations. Most such failures develop first from an anomaly into a critical state over time. These anomalies are usually reflected in the monitored signals, though not always visually distinguishable. Conventional approaches, which typically rely on such prominent changes in signals, therefore fail to identify failures beforehand. Identifying failure types at early stages, when they are still only anomalies in the signal, is of high importance. It allows for improved maintenance planning, minimizing time and revenue loss from operational disruptions. Leveraging deep neural networks, we propose our new methodology for predictive maintenance to classify anomalies sufficiently in advance into the type of failures in future. We demonstrate our method's efficacy on 10 failure cases for CVCM, independent of installation location or configurations. Our method is ISO-17359 compliant and especially successful in improving early-stage anomaly classification compared to conventional approaches, which is paramount for effective predictive maintenance. We achieve solid operational performance with promising results: 99.31\% overall classification accuracy across all failure types with early-stage detection averaging below 1\% into start of anomaly development in signals. Finally, utilizing conformal prediction techniques, we provide accompanying uncertainty quantification metric for maintainers to measure model confidence in a particular prediction, where we reach 99\% confidence, with sufficient coverage for each failure class. Given CVCM is deployed in various urban settings worldwide, our research holds significant relevance for maintenance personnel. Our methodology is scalable and can also be generalized across different track circuits and railway systems for improving reliability of railway operations through predictive maintenance.
\end{abstract}

\keywords{
    CVCM Track Circuits, Railway Predictive Maintenance, Anomaly Classification, Deep Learning, Conformal Prediction
}

\section{Introduction}

\subsection{Continuous Variable Current Modulation (CVCM) Track Circuit}

Track circuits are the predominant technology in use today for locating trains along the track. For each track segment, there is a transmitted electrical signal that travels through each rail and is then received on the other end. When a train passes, it shunts the electrical signal due to wheel contact, stopping reception. This indicates track occupancy for that segment, which is critical information for other trains to avoid possibility of collisions. CVCM \citep{schon:hal-00924934} is one of such track circuit technologies that has a transmitted signal (TX) from the centre of a track segment, going out into both directions. This results in 2 received signals (RX), namely CAT and CAL, in the upstream and downstream sections (Figure 1). Physically, the track is a continuous piece of hardware, so CVCM divides it into distinct, electrically isolated segments, by frequency of transmission and reception used. Frequencies typically range from around 8.2 – 11 kHz and adjacent segments always have different frequencies. Locating trains is imperative for operations, making track circuits safety critical. In the advent of any failures, the track circuit defaults to an occupied state, halting further operations.

\vspace*{-\baselineskip}
\vspace{0.3cm}
\begin{figure}[H]
    \centering
    \includegraphics[width=0.6\linewidth]{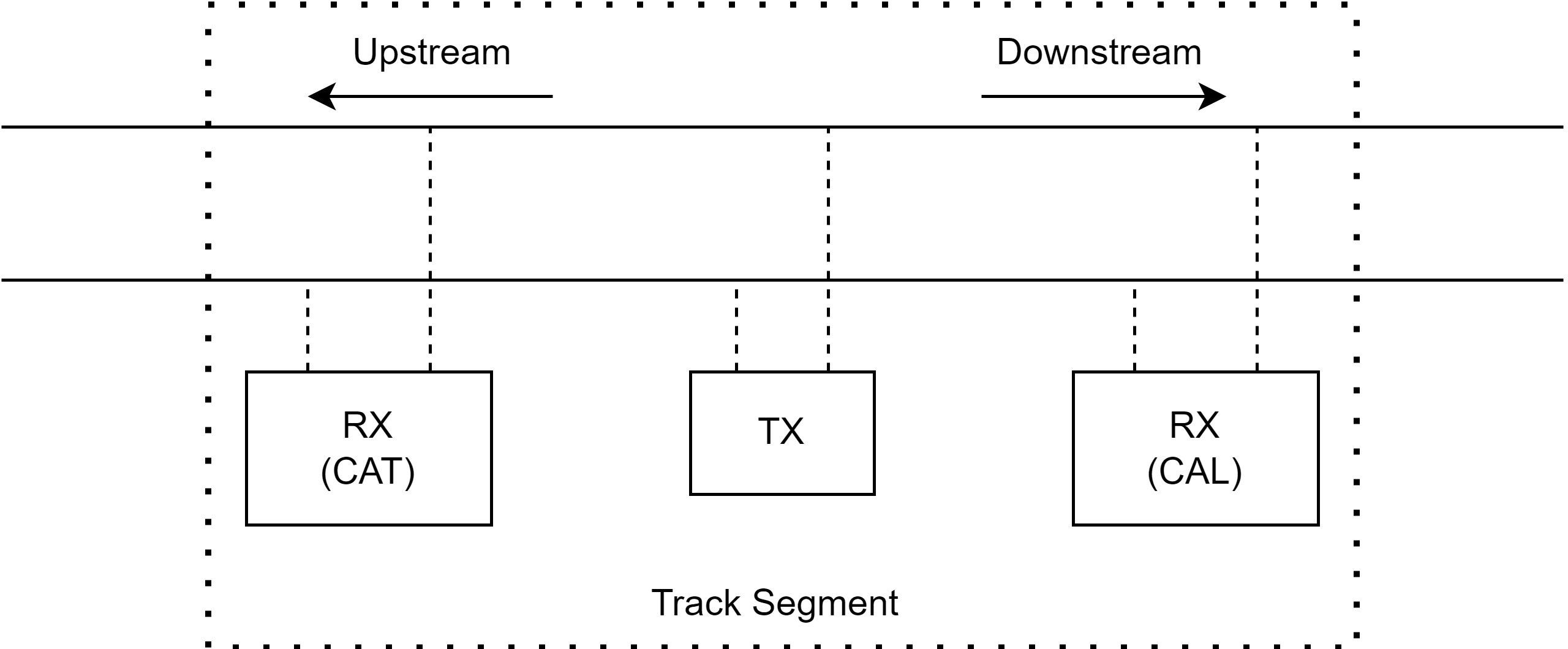}
    \caption{CVCM transmitted (TX) and received (RX) signals for a track segment}
    \label{fig:1}
\end{figure}
\vspace*{-\baselineskip}

\subsection{Problem Definition}

Failures in the CVCM hardware is therefore a significant problem faced by railway operators. The components typically undergo some degradation with use or due to unforeseen issues leading to different failures. This causes unwanted delays to trains and incurs costs such as revenue loss. Failures in the hardware are reflected in the characteristics of the received signals as they pass through the defective hardware components. The changes to the electrical signals are anomalous in nature from the usual nominal state of operation but could still be within operational tolerances, making them hard to distinguish. Additionally, a failure, before it reaches a critical state, will usually result from a progression of such anomalies that develop early on. Not all anomalies necessarily lead to an eventual failure so it is also necessary to discern this. It is vital to classify anomalies as early as possible during their initial formation to predict the failure types and time expected, so that maintenance can be planned in advance to minimize disruptions. Specifically for CVCM, due to its design, depending on which part (upstream or downstream), failures would affect each of the received signals differently.

\vspace*{-\baselineskip}
\vspace{0.4cm}
\begin{figure}[H]
    \centering
    \includegraphics[width=0.7\linewidth]{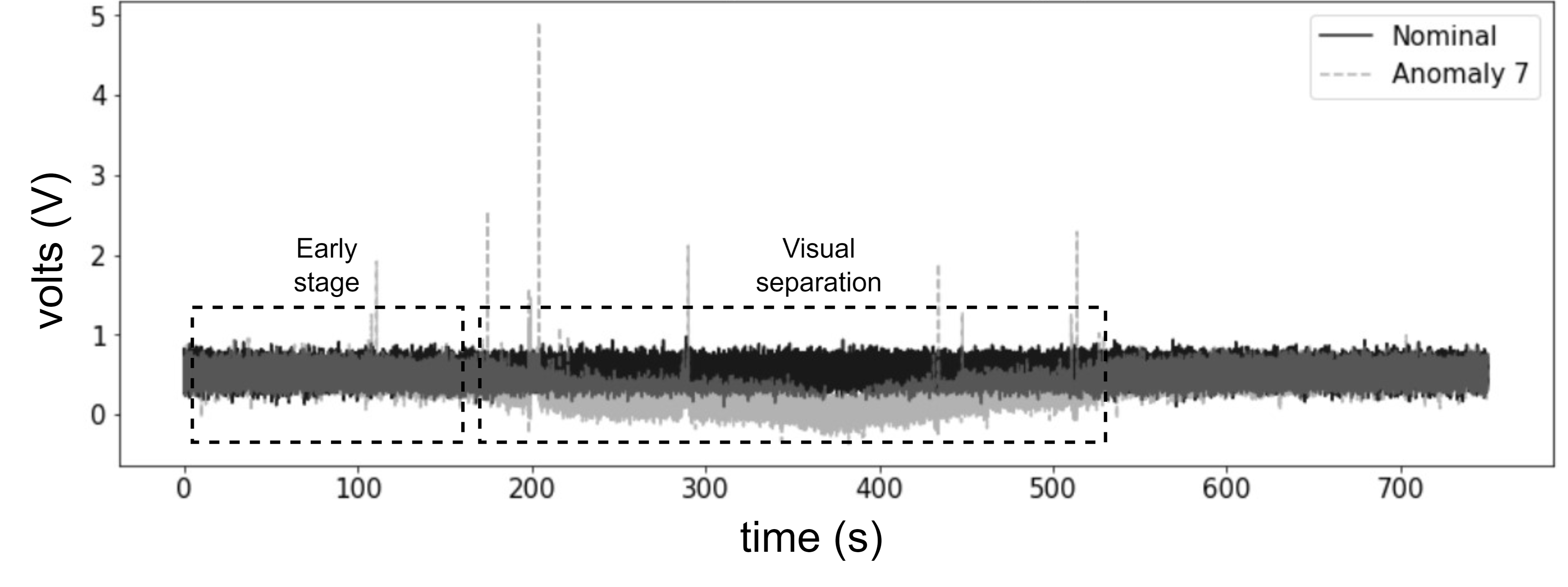}
    \caption{Example of 'Nominal' vs 'Anomaly' received signal (RX) profiles}
    \label{fig:2}
\end{figure}
\vspace*{-\baselineskip}
\vspace{0.2cm}

As an anomaly progressively degrades, compared to the nominal behaviour, there is a point in time where it deviates sufficiently to visibly tell the signal is unusual. However, at this stage the failure is already critical, so diagnosis relying on visual separation (Figure 2) cannot help to identify a failure before it happens. Early stage classification of anomalies, when it has not yet developed into a failure (when it is difficult to visually distinguish from nominal conditions) is a key business requirement for effective predictive maintenance that ensures safe and reliable operations.

\begin{figure}[H]
\begin{subfigure}{0.5\textwidth}
\hspace{0.3cm}
\includegraphics[width=1.0\linewidth]{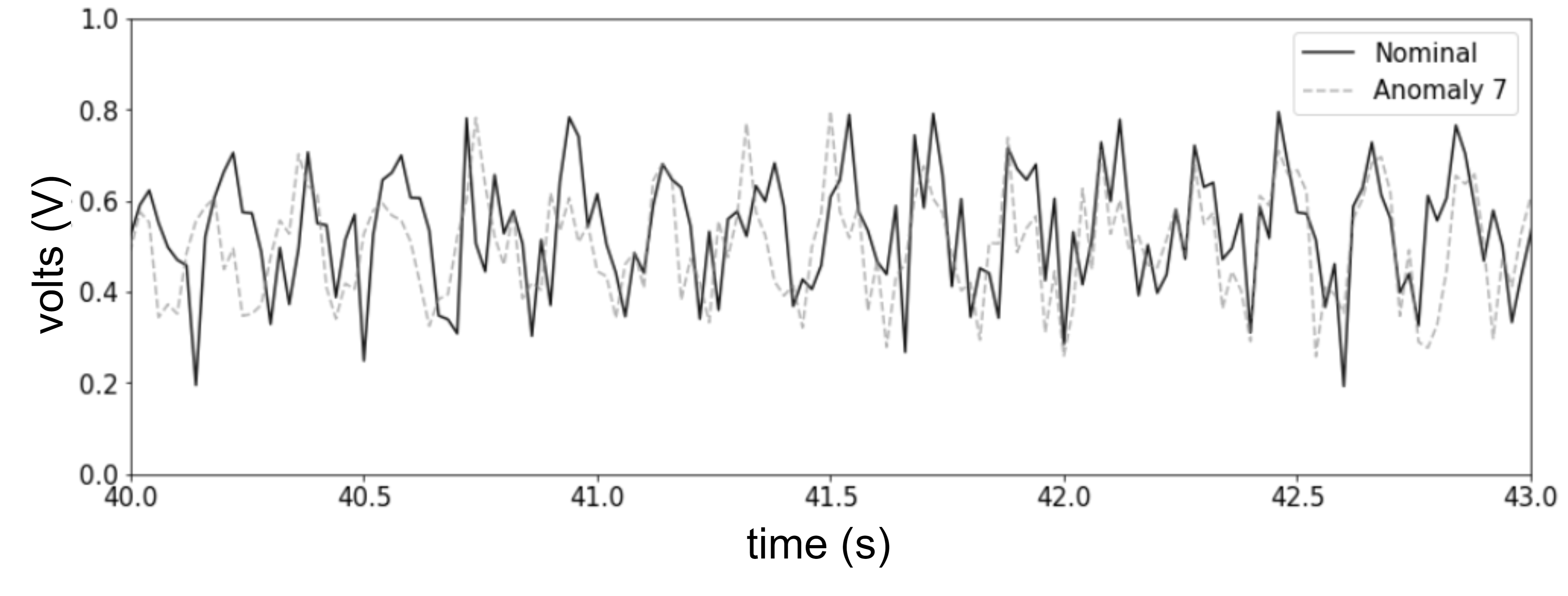} 
\caption{'Early Stage' nominal vs anomaly}
\label{fig:s1}
\end{subfigure}
\begin{subfigure}{0.5\textwidth}
\hspace{0.1cm}
\includegraphics[width=1.0\linewidth]{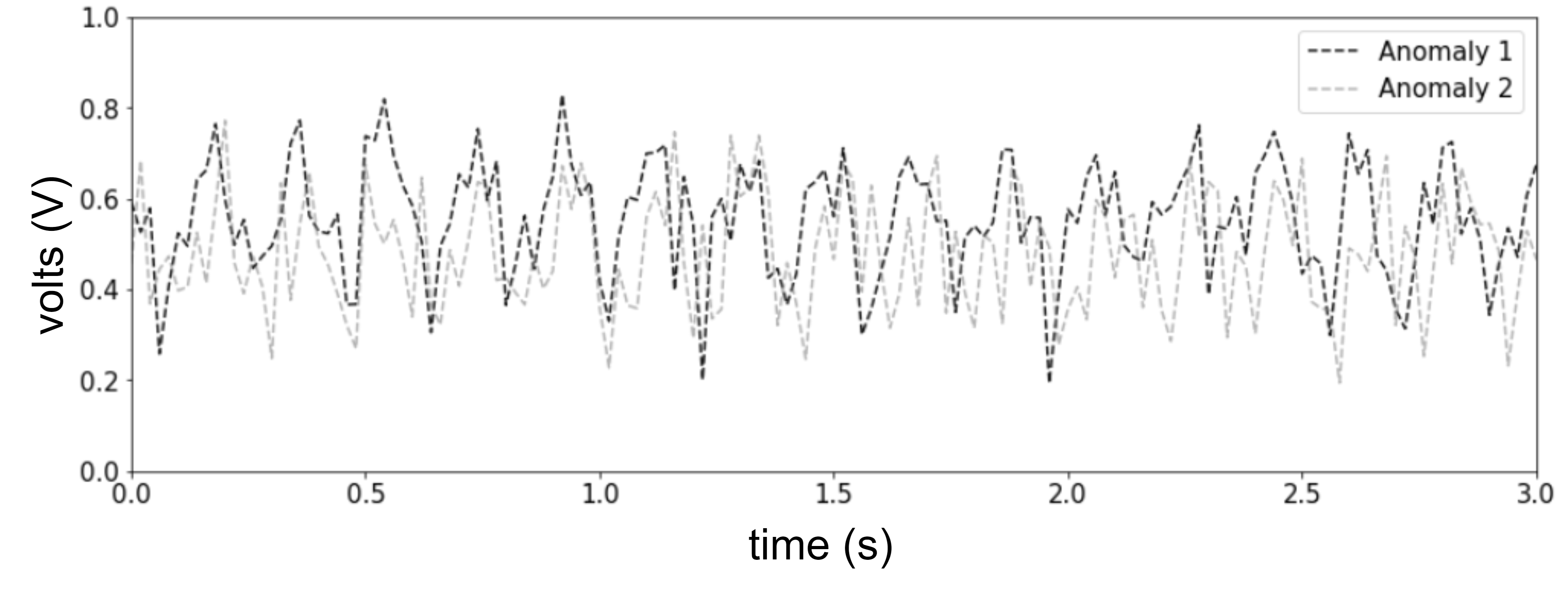}
\caption{'Early Stage' 2 different anomalies}
\label{fig:s2}
\end{subfigure}
\caption{CVCM 'Early Stage' signal comparisons}
\label{fig:3}
\end{figure}
\vspace*{-\baselineskip}

However, early stage anomalies look very similar to nominal signals (Figure 3(a)). Moreover, different anomalies are also not easily distinguishable between one another at these stages (Figure 3(b)). The main idea therefore is to successfully classify anomalies into possible failure types in future early enough (when changes may not yet be visible or evident to human eyes) before it has actually progressed into that failure, with high confidence. We propose our methodology (section 3), utilizing a set of deep neural network models to solve this problem.

\subsection{Use Cases and Data}

\subsubsection{Use Cases}

First we define the interested failure use cases for CVCM that require prediction at early stage. These use cases are based on operational experience and domain expert knowledge and are labelled as numbered anomaly classes for reference (in no particular order): 

\begin{table*}[ht]
    \centering
    \footnotesize
    \renewcommand{\arraystretch}{0.8}
    \begin{tabular}{|l||l|}
    \hline
         \textbf{Use case} & \textbf{Description}\\[1mm]\hline
         Anomaly 1 & Decrease of the ballast resistance simulated with a 1-ohm resistance\\[1mm]\hline
         Anomaly 2 &  Degradation of the LC downstream contact\\[1mm]\hline
         Anomaly 3 &  Degradation of the LC upstream contact\\[1mm]\hline
         Anomaly 4 &  Degradation of the track transformer contact\\[1mm]\hline
         Anomaly 5 &  Degradation of the wheel-rail contact\\[1mm]\hline
         Anomaly 6 &  Intermittent degradation of the railway bonding between the remote amplifier (RA) and the sensor\\[1mm]\hline
         Anomaly 7 &  Progressive degradation of the railway bonding between the remote amplifier (RA) and the receiver\\[1mm]\hline
         Anomaly 8 &  Intermittent degradation of the railway bonding between the remote amplifier (RA) and the receiver\\[1mm]\hline
        Anomaly 9 &  Progressive degradation of the railway bonding between the remote amplifier (RA) and the sensor\\[1mm]\hline
        Anomaly 10 &  Broken rail downstream\\[1mm]\hline
        Anomaly 11 &  Broken rail upstream\\[1mm]\hline
    \end{tabular}

    \caption{Anomaly use cases labels and descriptions}
    \label{tab:1}
\end{table*}
\vspace*{-\baselineskip}
\vspace*{-\baselineskip}

\subsubsection{Data}

In order to build and test our solution, we utilize signals data on the different failure use cases (Table 1) for CVCM. The proprietary dataset consists of 10 csv files containing the received signals of CVCM. There is one file per experiment of a failure use case produced in lab environment (except anomaly 6). It starts with fixed duration of nominal signal followed by the full anomaly profile before ending once again with a sequence of nominal signal. Thousands of pulses for each failure case was generated from its respective full signal profile to test with statistical significance on the performance in anomaly classification and determine earliness of predicting future failure types.

\section{Related Work}

\subsection{Conventional Approaches}

To establish a baseline, we look into conventional methods being used for track circuit failure diagnosis, typically involving visual inspections by maintenance specialists. As explained earlier, it is not possible to visually tell nominal and anomalies apart before failures get too close to critical. Automating the manual visual inspection process with feature extraction and signal analysis using wavelet transform \citep{8879833} have also been tested as alternatives. Additionally, other conventional approaches for condition monitoring \citep{6191870} involve simple thresholds over raw signal values or some signal characteristic such as mean or amplitude. If values fall outside defined nominal range, an issue is identified. Once again from our plots we observe how a simple threshold is unable to quantify the subtle changes and separate different anomalous signals easily at early stages. Moreover, the absolute values of signals change depending on location and configuration, so manual calibration of the thresholds for each segment configuration makes this approach not easily scalable. This justifies the need for a standardized methodology that can be applied for any track circuit to obtain reliable results. More importantly, it is key that the methodology has sufficient efficacy in preemptive classification of anomalies.

\subsection{Deep Learning Approaches}

Deep learning techniques are promising and demonstrate the potential to learn more complex feature representations from large datasets, leading to their gain in popularity for failure diagnosis \citep{8999070}. The most promising approach for industrial application was with successful usage for structural health monitoring \citep{inproceedings}, although in context of high frequency data (MHz range), which may not be practical for large scale operations in our context. With lower frequency, a study \citep{ryu2023development} found total success in 16 out of 19 events, but diving deeper we find it only worked well for 4 out of 19 events in early stages. Even though deep learning techniques can be successful, early identification still proves to be a challenge.

There is work done specifically for unlabelled track circuit signals \citep{10.1115/JRC2019-1300} to cluster pulses based on similarity of shapes. This is especially useful for field data that may not have labels for the different conditions so once the clustering is done, further analysis on each group of signal pulses can be done to identity the different anomalous conditions. After expert validation, the logical next step is to then learn from such data so that for new signals coming in, the anomalies can be separated into the respective probable future failure types. For this, a new methodology is necessary.

A review conducted on deep learning based approaches for predictive maintenance \citep{article} also does not show any established comprehensive methodology for track circuits that solves early-stage classification and prediction with high confidence. This motivated the need for developing a generalized methodology leveraging deep learning techniques, that can easily be tailored for different use cases and provide useful results in failure diagnostics for predictive maintenance.

\section{Methodology}

Implementing an effective strategy for predictive maintenance requires understanding that failures in systems using electrical signals have a precursor impact reflected in these signals prior to failure occurrence. Before any failures emerge in the system, it typically has its roots manifested as some form of an anomaly that starts to develop in the signals. Anomalies would be conditions where the signal characteristics deviate from what is typically nominal but might still be within acceptable operating limits, making them undetectable using conventional diagnosis methods examined previously. As anomalies start to become more pronounced, they eventually deviate sufficiently from normal functioning of a particular system, exceeding those limits, resulting in identifying it during or close to critical failure state. With this understanding, we develop a robust and adaptable methodology for working with any single or multi-channel time series electrical signals, found in numerous rail infrastructure systems, such as track circuits. The goal is predicting well in advance what type of failure is expected to happen in the future with sufficient reliability to facilitate effective predictive maintenance. Distinguishing the subtle changes to the signals is necessary at the stage where anomalies are starting to develop early on and have not yet progressed into where the system reaches the respective distinct failure states. To achieve this, our methodology is defined as follows:

\begin{figure}[H]
    \centering
    \includegraphics[width=0.8\linewidth]{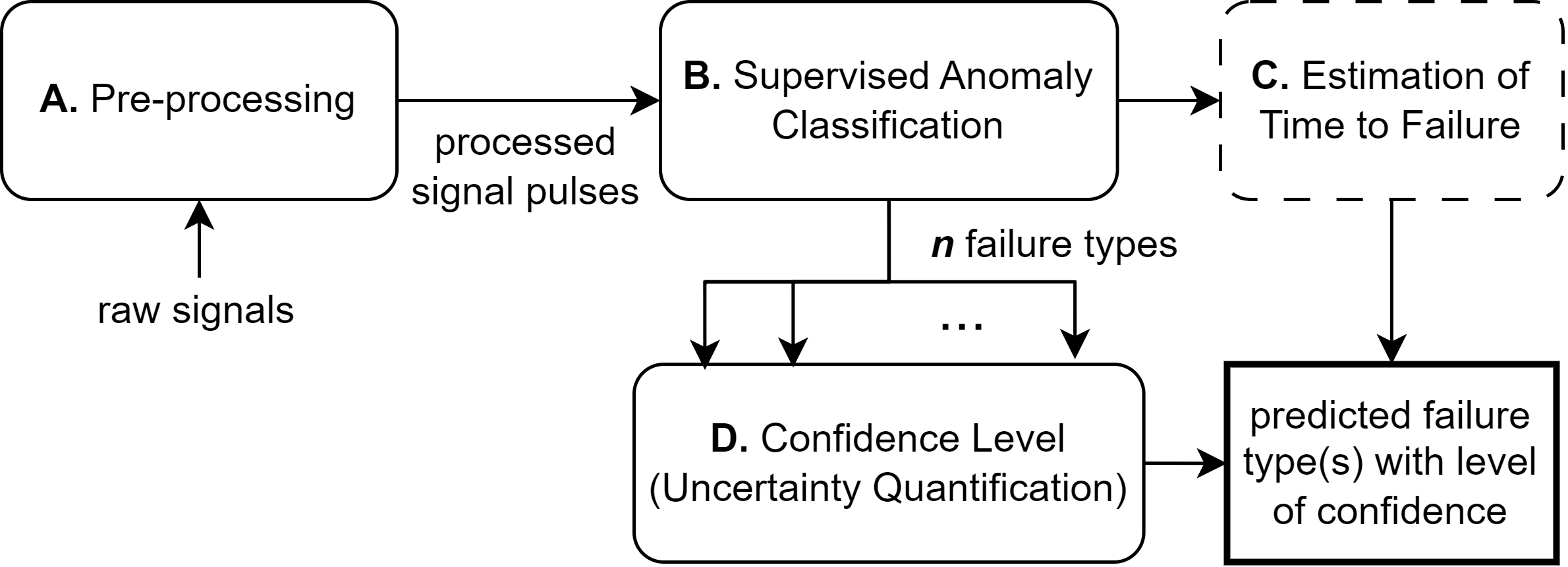}
    \caption{Methodology for pre-emptive failure diagnostics in track circuits}
    \label{fig:4}
\end{figure}
\vspace*{-\baselineskip}
\vspace{0.2cm}

First, pre-processing of the raw signals using a combination of signal processing, statistical and analytical techniques are conducted to remove the effects of noise perturbations and transform them into a form suitable for subsequent modelling steps, while preserving its key characteristics. Next, a deep supervised anomaly classifier learns the patterns of each type of anomaly from its full signal profile so that it can identify the failure type at any stage, especially early on. It is then possible to create an approximator, that based on which part of the anomaly profile is being detected, estimates time to critical failure. Lastly, confidence level in a particular prediction can also be quantified. In reality, presence of inherent noise or randomness (aleatoric uncertainties) plus limitations in how much the sampled data is representative of the population space (epistemic uncertainties) means that models which typically give point predictions fail to account for these variations or uncertainties \citep{gawlikowski2023survey} leading to inaccuracies in the outputs. A more useful set of results can quantify these uncertainties in model predictions and hence determine the most likely set of values for practical decision-making, where the desired confidence can be achieved. Our methodology has proved to be both generalizable and scalable. Additionally it is also compliant with the ISO-17359 \citep{iso} specification.

\section{Results}

\subsection{Anomaly Classification}

We quantify the performance of our deep supervised anomaly classifier by calculating the overall classification accuracy across all 10 use cases with uniform distribution of classes and across all stages of the anomalies formation. The classification achieves an overall 99.33\% accuracy. We also visualize the performance for each individual use case from the confusion matrix, which is a plot of each of the true class labels against the predicted class labels to give a grid of the number of predictions in each instance. In a theoretically ideal scenario, using a normalized confusion matrix plot, where the values are scaled between 0 and 1, we want the diagonal to all be 1 in value and 0 elsewhere. The confusion matrix (Figure 5) summarizes the results of our deep supervised anomaly classifier model in greater detail.

\begin{figure}[H]
    \centering
    \includegraphics[width=0.6\linewidth]{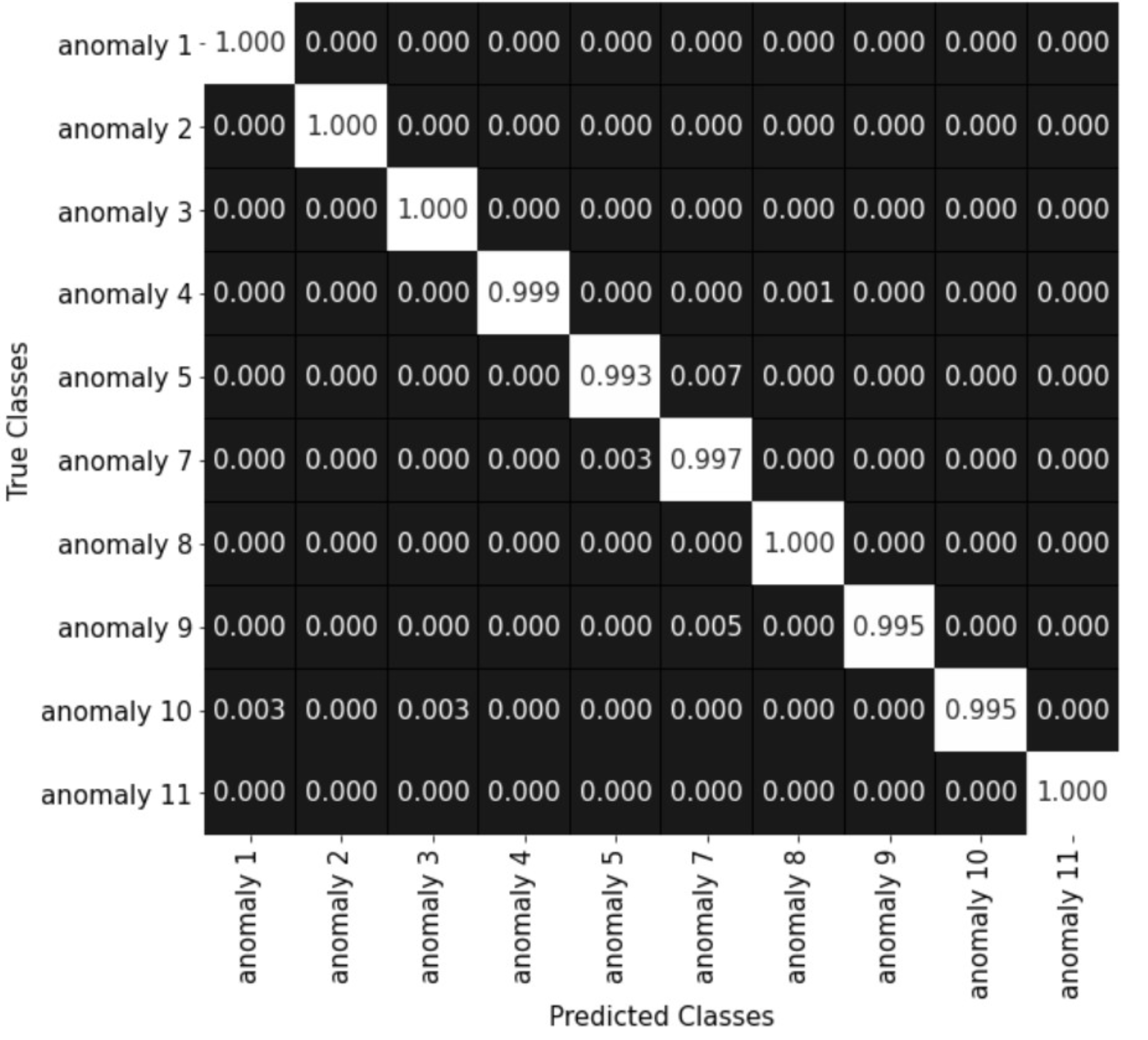}
    \caption{Confusion Matrix (Normalized) - overall classification performance across 10 failure categories}
    \label{fig:5}
\end{figure}
\vspace*{-\baselineskip}
\vspace{0.2cm}

As evident, our results are very close to ideal. Since the anomaly pulses tested were across the full signal profile, classification to identify failure types in early stage for each case was also very successful. This means our methodology is able to separate anomalous signals despite only subtle differences early on and predict the different failure types in future.

Conventionally, after the full or substantial part of the failure signal is observed, the diagnosis is possible on what failure it is, but at that point it is too late. The benefit of our method is that we do not require the full anomaly signal profile in order to determine what type of failure it is. In fact, with only a small pulse window and that too from any part of the signal profile our model is able classify with high accuracy as it learns to map it to the respective failure cases. This is why conventional diagnostics is done after the fact and is more reactive, while our method not only automates the process of identification but it also achieves it well in advance.

For the very small number of confused cases, meaning the instances where a signal pulse might be confused to be a different failure and therefore misclassified, we can address this better through the use of conformal prediction to guarantee a certain level of confidence in our prediction sets (section 4.3).

\subsection{Earliness of Classification (possible time to failure estimation)}

\vspace*{-\baselineskip}
\begin{figure}[H]
\begin{subfigure}{0.5\textwidth}
\hspace{0.1cm}
\includegraphics[width=1.0\linewidth]{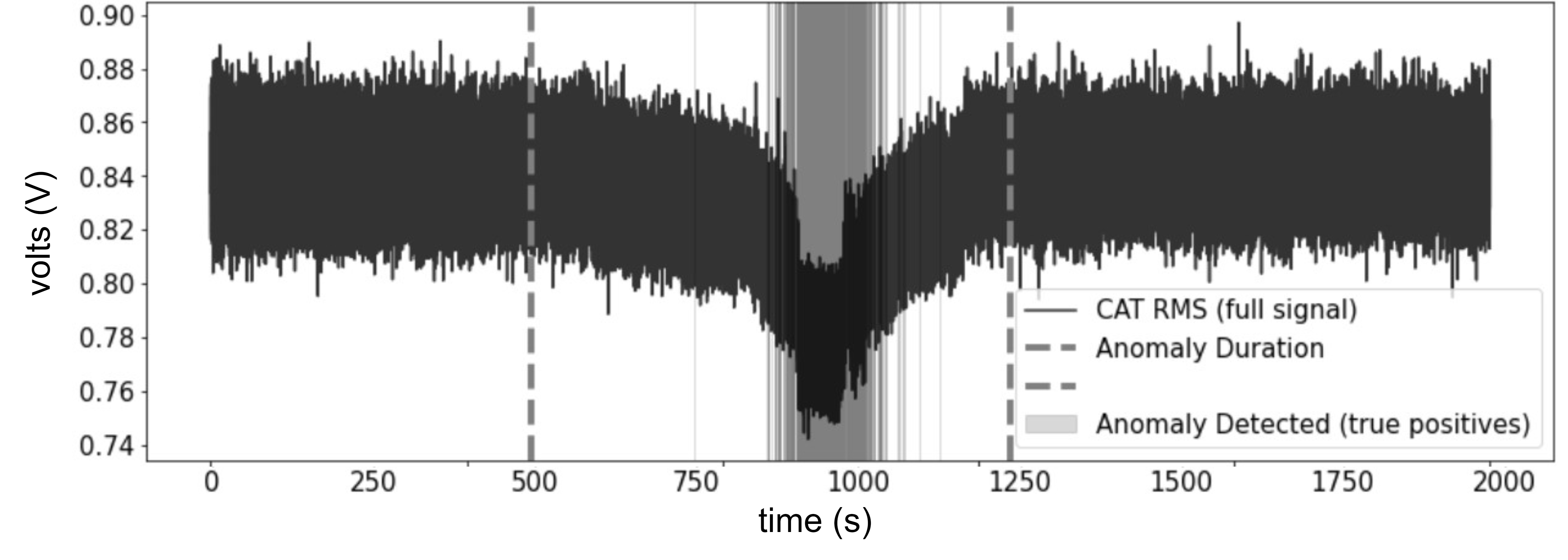} 
\caption{Conventional Method}
\label{fig:s9}
\end{subfigure}
\begin{subfigure}{0.5\textwidth}
\includegraphics[width=1.0\linewidth]{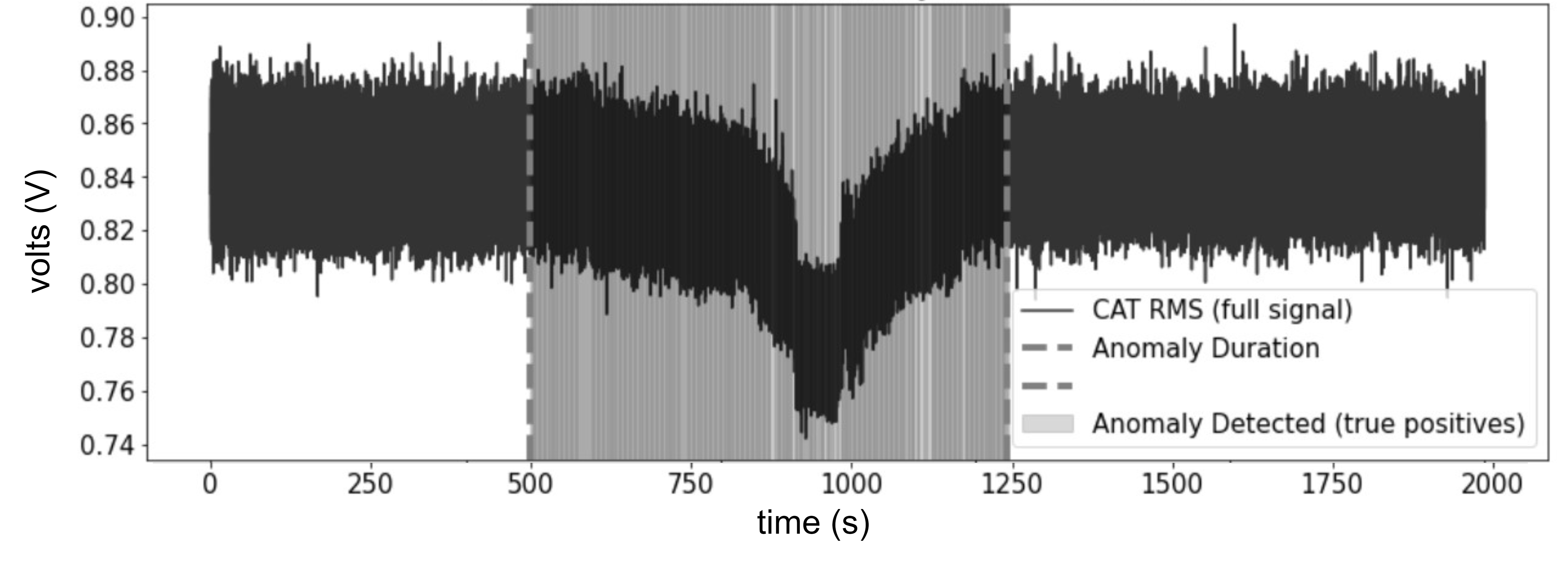}
\caption{Our Method}
\label{fig:s10}
\end{subfigure}
\caption{Differences in earliness of an anomaly classified into failure type (conventional vs ours)}
\label{fig:6}
\end{figure}

Another important success factor to quantify is the earliness of anomaly classification possible, due to its importance in effective predictive maintenance. We can illustrate this using an example anomaly signal profile. We locate the moment when the first anomalous pulse is successfully classified into its associated failure type for both conventional and our method. We then compute the percentage of time into when the first anomaly pulse is successfully classified from its actual start of development in the signal, to point of absolute critical failure. The times are normalized and expressed here in percentages, as each failure case could develop over its own time frame, depending on multiple factors (location, track usage, etc.), but the principle effect persists.

In our example (Figure 6), we observe the conventional approach classifies the upcoming failure at 58.95\% of elapsed time from when the anomaly first started to develop and to when the critical failure actually happens, so over halfway into its progression. Our method helps to reduce this time to 0.83\% from the start of the anomaly development, meaning more than 99\% of the time in advance before the most critical point of failure. This is an important metric, especially since early classification of type of failure is very difficult conventionally as early stage signals look very similar. However, with our deep learning based model, we can differentiate between failure types from early pulses well in advance as required.

From this early classification of anomalies it is also possible to expand into better time to failure estimations through approximating the development of anomalies with time for each case, which we aim to pursue as future work.

It is also important to note here that from when the first pulse is classified, as the anomaly progresses and our classifier keeps identifying the same anomaly, we can have higher confidence in the accuracy of our predicted failure type. This notion of confidence in prediction is important to have as part of our results for practical applications as it will help in making better informed decisions.

\subsection{Confidence Level}

A measure of confidence on the associated model outputs is key for operational decision-making. Typically, in the real world there are presence of different uncertainties be it due to noise or incomplete information that causes predictions to have some likelihood of varying from the ground truth. This means if a point prediction, or in other words a singular value is given as the model output, it fails to account for these variations leading to lower confidence on the accuracy of the predictions. A good practice is to quantify these uncertainties such that a measure of confidence can be achieved on top of the model predictions.

Conformal prediction \citep{angelopoulos2022gentleintroductionconformalprediction} is a reliable technique for quantifying uncertainties in modelling and hence measure how well model predictions conform to ground truth. The conformal methodology can be applied posterior to model training, on data that is independent and identically distributed (i.i.d). For example, we apply this technique for our classification model that predicts the failure classes. First, the acceptable level of uncertainty in the model prediction, $\alpha$, is defined e.g. 1\%. Conversely, there is 1-$\alpha$ (99\%) confidence that the prediction set will contain the true value. A prediction set means the set of most likely class labels that guarantee the desired confidence level. Prediction sets with only 1 label means it is highly confident that the prediction matches ground truth and for sets more than one, it is useful in narrowing down the most possible failure types. This is very useful for cases more similar to each other, where there is higher risk of getting an incorrect class prediction if only one label (point prediction) is given without any objective confidence associated and hence a possibility of missing the true class. Instead, giving say 2 most possible failure classes with a 99\% confidence is much more useful in diagnosis and maintenance planning. This is how possible confusions in classification can be mitigated, as mentioned about in section 4.1.

We utilize conformal prediction techniques and implement it as part of our methodology to provide statistically reliable uncertainty quantification such that we achieve high confidence on our model outputs.

\begin{figure}[H]
    \centering
    \includegraphics[width=0.7\linewidth]{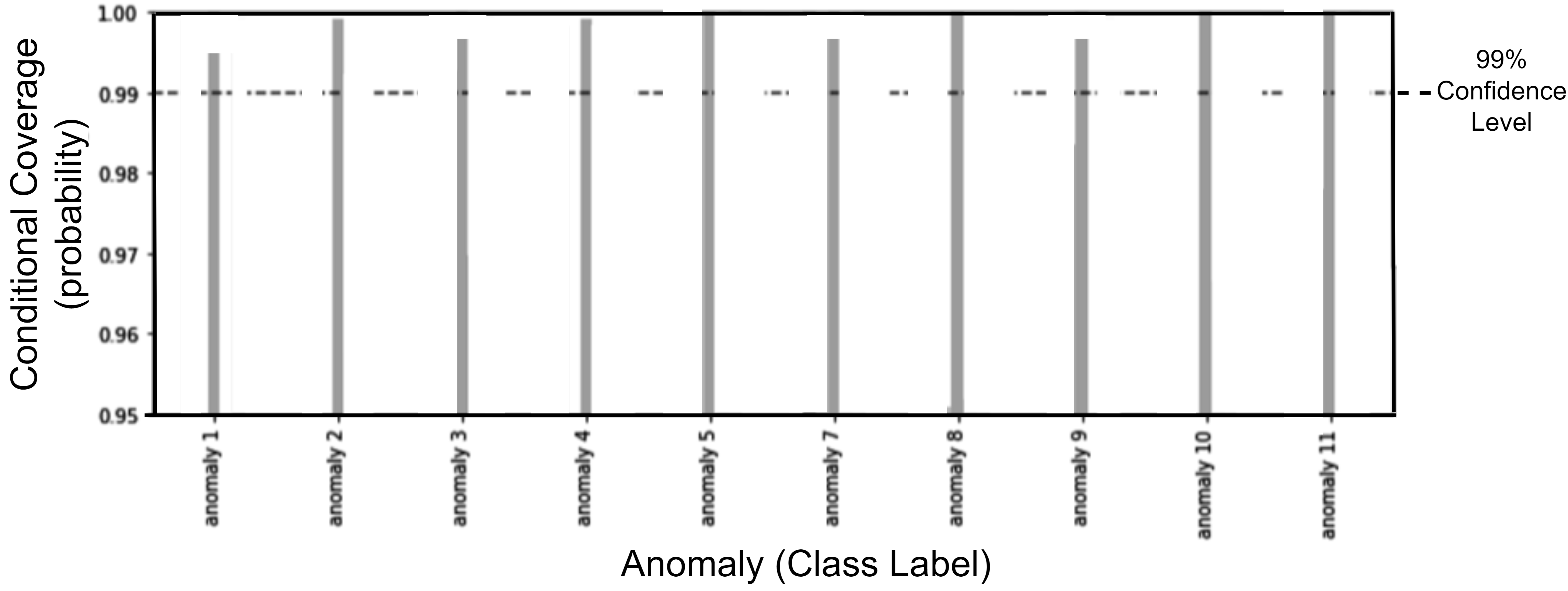}
    \caption{Conditional coverage for each class compared to 99\% confidence level}
    \label{fig:18}
\end{figure}
\vspace*{-\baselineskip}
\vspace{0.2cm}

We demonstrate effective uncertainty quantification and its utility with our anomaly classifier outputs. We define an acceptable risk level of uncertainties in the prediction as 1\%, meaning our desired confidence is 99\%. This implies there is 99\% probability that the derived prediction set of likely failure types contains the true label, i.e. guarantees coverage at that level. This increases both reliability and practicality of model predictions by providing not just a value but the best set of possible outcomes with an objective measure of confidence. This is especially useful for early stage diagnosis where pulses are more similar, so likely more uncertainty in predicting the right class between anomalies. A standard point prediction (single class label) will more easily be wrong on average as the likelihood of getting the right class is lower. Therefore being able to derive prediction sets with the most likely classes that guarantees coverage at a high confidence level is more useful and helps with false positives and false negatives. This increases the trustworthiness of the system. We were able to achieve this sufficient coverage across all 10 failure class predictions (Figure 7) at 99\% confidence level. However too many labels in the prediction sets in order to meet this coverage level will also not be useful so we look at the average prediction set size that we achieve with our model, which was found to be 1.06 (ideal is 1) meaning almost always there is only 1 class label predicted, which is 99\% likely to be ground truth. This means model precision is maintained while having a high confidence. Another advantage is detecting unknown anomalies, meaning if there are other anomalies that the classifier cannot assign any of the labels with high enough confidence then it returns a null set indicating the presence of a new unknown anomaly. Lastly, if there is a mixture of anomalies it helps by giving a probability of every fault for consideration. Better understanding of how an anomaly could relate to possible failure types allows for better informed decision-making.

\section{Discussion}

Misclassifications can be costly in terms of wrong failure diagnosis that would reduce the efficacy of predictive maintenance strategies. Our model was successful in minimizing possible misclassifications with a high overall as well as per failure classification accuracy. Moreover, conformal prediction was a useful layer on top of the model outputs in further mitigating possible errors and giving useful information for decision-making and pre-emptive diagnostics.

Another point of further investigation is length of pulse window relative to the anomaly profile required for classification. In general, longer windows would provide more accurate classification with a higher confidence but it also means it takes longer to determine the failure types, meaning later into start of the anomaly progression, as operators would have to wait for enough data to be logged for an accurate determination. The acceptable trade-off between pulse length and earliness of classification with acceptable accuracy can be further explored based on operational requirements in level of early detection required, which would have to be determined based on use cases. Having overlapping sliding windows as we implemented, helps reduce effects of noise or uncertainties in pulse regions while also reducing overall length of signal required to obtain results.

Lastly, the full progression of anomalies obtained from simulated failure experiments are assumed to be linearly translatable in reality. For example, an anomaly being created over minutes in the lab could take months to happen in reality, and different parts of the progression could occur over variable time frames based on additional factors present in the field. While this should not affect accuracy of classification of the failure directly, it could impact the duration taken and estimation of time remaining to critical failure, which we have mentioned would be part of our future work planned. Further investigation on the rate at which each failure evolves should be performed with validation in the field.

\section{Conclusion}

To summarize, our deep learning based methodology when applied to CVCM achieves early stage anomaly classification into 10 different failure cases sufficiently in advance and with high confidence. The subject-relevant expert validates our findings and confirms our solution hugely improves existing maintenance procedures, benefiting operations and maintenance activities greatly through automation. The key differentiating factor being the ability to predict in advance, thereby minimizing delays and disruptions through planned maintenance rather than react when failures occur. This is of tremendous value to operators. Moreover, there is no need for new sensors or data but instead we leverage already existing signals from the track circuits, making it easy to add to existing infrastructure as an algorithmic predictive maintenance solution over existing technologies. Our promising results establish our solution as scalable for large operational requirements and also generalizable for diagnostic and predictive maintenance for track circuits and beyond to various railway systems utilising time series electrical signals. Lastly, a very important consideration is the compliance of our novel method with the standards and safety practices of non-interfering solutions. The ISO 17359 provides guidelines for condition monitoring and diagnostics of machines and our proposed methodology complies with the requirements, though a further external audit should be conducted to validate this.

\section*{Author Contributions}
{Debdeep Mukherjee and Di Santi Eduardo contributed equally to this work as co-first authors.}

\bibliographystyle{unsrt}
\bibliography{references}

\end{document}